\documentclass[conference]{IEEEtran}

\usepackage{graphicx} 
\usepackage{tcolorbox}
\usepackage{tabularx} 
\usepackage{booktabs}
\usepackage{xurl}

\usepackage{array}
\usepackage{makecell}

\usepackage[colorlinks=true, linkcolor=blue, citecolor=blue]{hyperref}

\providecommand{\keywords}[1]
{
  \small	
  \textbf{\textit{Keywords---}} #1
}

\title{From Convolution to Transformer: A Comparative Study of U-Net Variants for Brain Tumor and Retinal Vessel Segmentation}

\author{
Khoa Pham\textsuperscript{1}, 
Sindhuja Penchala\textsuperscript{2},
Jiacheng Li\textsuperscript{2},
Andy Perkins\textsuperscript{1},
Noorbakhsh Amiri Golilarz\textsuperscript{2}\\
\textsuperscript{1}Mississippi State University,
\textsuperscript{2}The University of Alabama}

\begin{document}

\maketitle

\begin{abstract}
Medical image segmentation plays an important role in computer aided diagnosis, treatment planning, and disease monitoring. U-Net has been widely used for biomedical image segmentation because of its encoder decoder structure and skip connections. However, conventional convolution based U-Net models may have limited ability to capture long range dependencies and global contextual information, which can affect performance in complex segmentation tasks. This paper presents a comparative study of five U-Net based architectures: U-Net 3D, Residual U-Net, Attention U-Net, UNETR, and Swin UNETR. The models are evaluated on two benchmark datasets: BraTS 2023 for brain tumor segmentation and DRIVE for retinal vessel segmentation. Experimental results show that Swin UNETR achieves the best overall performance, with Dice scores of 0.8965 on BraTS 2023 and 0.8078 on DRIVE. The results suggest that transformer based U-Net variants are effective for segmentation tasks requiring global contextual modeling, while residual learning remains useful for fine structure segmentation. This study provides practical insights into model selection for medical image segmentation across volumetric MRI and retinal imaging tasks.
\end{abstract}

\keywords{Medical Image Segmentation, Brain Tumor, 3D U-Net, Residual U-Net, Attention U-Net, UNETR, Swin UNETR}

\section{Introduction}
Medical image segmentation is a fundamental task in computer aided diagnosis because it enables the identification and delineation of anatomical and pathological structures, such as tumors, organs, and blood vessels \cite{moghbel2018review}. Accurate segmentation can support clinical diagnosis, treatment planning, surgical guidance, disease monitoring, and quantitative image analysis \cite{raposo2024intelligent}. Traditionally, medical image segmentation has relied heavily on manual annotation by medical experts. However, manual segmentation is time consuming, labor intensive, and subject to inter observer variability. These limitations have motivated the development of automated segmentation methods based on deep learning.

Among deep learning architectures, U-Net has become one of the most widely used models for biomedical image segmentation \cite{azad2024medical}. Its encoder decoder structure captures semantic information through downsampling and restores spatial resolution through upsampling. Skip connections further allow low level spatial features to be combined with high level semantic representations, making U-Net effective for pixel level and voxel level prediction \cite{seo2019modified}. However, the baseline U-Net mainly relies on local convolutional operations, which can limit its ability to capture long range dependencies and global contextual relationships \cite{azad2024medical}. This limitation becomes more important in complex medical imaging tasks, such as brain tumor segmentation, where tumor regions may have irregular shapes, heterogeneous appearances, and ambiguous boundaries \cite{abueed2025systematic}.

To address these limitations, several U-Net variants have been proposed. Residual U-Net improves feature propagation and training stability through residual learning \cite{alom2019recurrent}. Attention U-Net introduces attention gates to emphasize task relevant regions and suppress less useful background information \cite{schlemper2019attention}. Transformer based architectures, such as UNETR \cite{hatamizadeh2022unetr} and Swin UNETR \cite{hatamizadeh2021swin}, further improve global context modeling by using self attention mechanisms. These variants provide different advantages, but their performance may vary across imaging tasks, data dimensionality, and target structure. Therefore, a comparative evaluation is needed to better understand how different U-Net based architectures perform under different medical segmentation scenarios.

In this study, we compare five representative U-Net based models: U-Net 3D, Residual U-Net, Attention U-Net, UNETR, and Swin UNETR. The evaluation is conducted on two benchmark medical image datasets: BraTS 2023 for brain tumor segmentation \cite{brats2023} and DRIVE for retinal vessel segmentation \cite{staal2004drive}. These datasets represent different segmentation challenges. BraTS 2023 involves volumetric multimodal MRI segmentation with complex tumor subregions, while DRIVE involves two dimensional fundus image segmentation with thin and branching vessel structures. By evaluating the models on both datasets, this study examines the effectiveness of convolution based, residual, attention based, and transformer based U-Net variants across different medical imaging tasks.

The main contributions of this paper are summarized as follows: First, this paper provides a comparative evaluation of representative U-Net variants on both brain MRI and retinal imaging datasets. Second, it analyzes how different architectural mechanisms, including residual learning, attention gates, and transformer based self attention, affect segmentation performance. Third, it discusses practical model selection considerations for volumetric tumor segmentation and fine structure vessel segmentation.

The remainder of this paper is organized as follows. 
Section \ref{Methodology} describes the architectural characteristics of the evaluated models. Section \ref{Performance Evaluation} presents the performance evaluation, including datasets, preprocessing, quantitative results, and comparative discussion. Section \ref{Conclusion} concludes the paper and outlines future research directions.

\section{Methodology}
\label{Methodology}

This section presents the U-Net based architectures evaluated in this study: 3D U-Net, Residual U-Net, Attention U-Net, UNETR, and Swin UNETR. These models represent major segmentation designs, including convolutional encoder decoder learning, residual learning, attention based feature selection, and transformer based global context modeling.

\subsection{Overview of Compared U-Net Architectures}
U-Net has become one of the most widely used architectures for medical image segmentation because of its encoder decoder structure and skip connections \cite{azad2024medical}. The encoder gradually extracts high level semantic features by reducing spatial resolution, while the decoder restores spatial resolution to generate dense segmentation masks \cite{siddique2021u}. Skip connections transfer fine grained spatial information from the encoder to the decoder, allowing the model to combine localization details with semantic representations \cite{du2020medical}.

Although the original U-Net is effective for many biomedical segmentation tasks, its convolutional operations mainly focus on local spatial patterns \cite{ronneberger2015u}. This limitation can reduce performance when the target structures have irregular shapes, weak boundaries, or long range contextual relationships. Therefore, several U-Net variants have been developed to improve feature learning, localization accuracy, and contextual modeling. Residual U-Net improves gradient propagation through residual learning \cite{alom2019recurrent}. Attention U-Net uses attention gates to emphasize task relevant regions \cite{schlemper2019attention}. UNETR introduces transformer based encoding to capture long range dependencies \cite{hatamizadeh2022unetr}. Swin UNETR further improves transformer based segmentation through hierarchical representation learning and shifted window self attention \cite{hatamizadeh2021swin}.
\vspace{-0.2cm}
\subsection{3D U-Net}
3D U-Net extends the original two dimensional U-Net architecture to volumetric medical image segmentation \cite{mehta20183d}. Instead of using two dimensional convolution, pooling, and upsampling operations, 3D U-Net applies three dimensional operations throughout the encoder and decoder. This design allows the model to learn spatial relationships not only within each image slice but also across adjacent slices.

The encoder of 3D U-Net progressively extracts volumetric feature representations through 3D convolutional layers and downsampling operations \cite{isensee2019attempt}. As the spatial resolution decreases, the network captures increasingly abstract semantic information. The decoder then reconstructs the segmentation output through 3D upsampling operations \cite{tie2021mri}. Skip connections between corresponding encoder and decoder stages preserve spatial details that may be lost during downsampling \cite{mao2016image}.

This architecture is particularly suitable for brain MRI segmentation because MRI data are naturally volumetric. By modeling inter slice information, 3D U-Net can capture anatomical continuity across slices and provide a stronger baseline than a purely two dimensional model for volumetric segmentation tasks. However, because 3D U-Net mainly relies on local convolutional kernels, its ability to capture long range dependencies remains limited compared with transformer based models \cite{jia2022u}.

\subsection{Residual U-Net}
Residual U-Net incorporates residual learning into the U-Net framework \cite{alom2019recurrent}. In deep neural networks, increasing the network depth can improve representation capacity, but it may also introduce optimization difficulties, such as vanishing gradients and degraded training performance. Residual connections address this problem by allowing feature information to bypass one or more convolutional layers and flow directly to deeper layers \cite{li2019residual}.

\begin{figure*}[!t]
\centering
\includegraphics[width=0.8\textwidth]{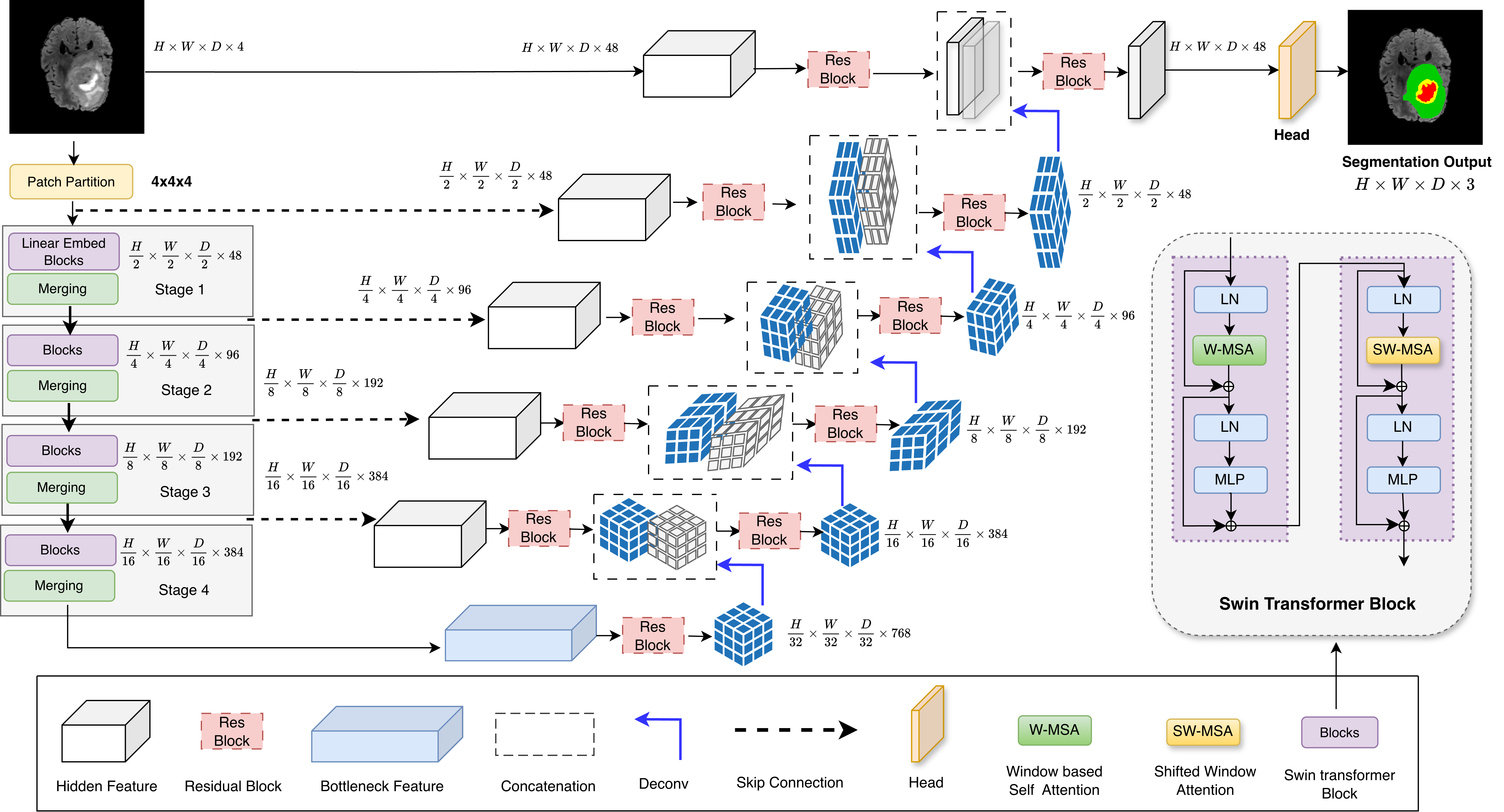}
\caption{Swin UNETR model for 3D brain tumor segmentation \cite{11076674} \cite{hatamizadeh2021swin}, integrating Swin Transformer encoder blocks with residual convolutional decoder blocks to capture both global contextual information and local spatial features. Window based and shifted window self attention mechanisms are used to learn hierarchical representations efficiently, while skip connections and deconvolution layers reconstruct high resolution segmentation outputs.}
\label{fig:BTSSwinUNETR}
\end{figure*}

In Residual U-Net, standard convolutional blocks are replaced or enhanced by residual blocks \cite{alom2019recurrent}. Each residual block learns a residual mapping instead of directly learning the entire feature transformation \cite{alom2019recurrent}. This design improves gradient propagation and makes it easier to train deeper segmentation networks. As a result, the model can extract more complex features while maintaining training stability.

For medical image segmentation, residual learning is useful because anatomical and pathological structures can have complex shapes and heterogeneous appearances. In brain tumor segmentation, residual blocks can help the model learn richer tumor representations \cite{shehab2021efficient}. In retinal vessel segmentation, residual learning can support the extraction of thin and branching vessel patterns \cite{chen2021retinal}. However, Residual U-Net still depends mainly on convolution based feature extraction, which means that its global contextual modeling capability is limited by the receptive field of the convolutional operations.

\subsection{Attention U-Net}
Attention U-Net introduces attention gates into the U-Net architecture to improve the selection of task relevant features \cite{schlemper2019attention}. In the original U-Net, encoder features are directly transferred to the decoder through skip connections \cite{ronneberger2015u}. Although these skip connections preserve spatial details, they may also transfer irrelevant background information. Attention gates address this issue by assigning higher weights to important regions and lower weights to less relevant regions before feature fusion \cite{schlemper2019attention}.

The attention mechanism allows the model to focus on anatomical or pathological structures that are more important for segmentation. For brain tumor segmentation \cite{khan2020brain}, attention gates can help highlight tumor regions and suppress normal brain tissues or background areas \cite{nodirov2022attention}. For retinal vessel segmentation, attention gates can help emphasize thin and low contrast vessels while reducing the influence of noise \cite{golilarz2017translation} \cite{golilarz2018hyper} and nonvessel structures \cite{guo2021sa}.

Attention U-Net is especially useful when the target region occupies a small portion of the image or has weak boundaries. By selectively enhancing important features, the model can improve localization accuracy and reduce false positive predictions. However, attention gates in Attention U-Net are usually added to a convolution based framework, so the model may still have limited ability to capture long range dependencies compared with transformer based architectures \cite{azad2022contextual}.

\subsection{UNETR}
UNETR introduces a transformer based encoder into the U-Net style segmentation framework \cite{hatamizadeh2022unetr}. Unlike traditional U-Net models that use convolutional encoders, UNETR represents the input volume as a sequence of image patches and processes these patches using a transformer encoder. This design allows the model to capture long range dependencies and global contextual relationships across the input volume \cite{hatamizadeh2022unetr}.

In UNETR, the input medical volume is divided into nonoverlapping patches \cite{shaker2024unetr++}. These patches are converted into embedded token representations and passed through transformer layers. The self attention mechanism allows each token to interact with other tokens, enabling the model to learn global relationships that may not be captured effectively by local convolutional kernels. The transformer encoder produces hierarchical feature representations that are connected to a convolutional decoder through skip connections.

The decoder of UNETR follows the general U-Net design and progressively reconstructs the segmentation mask \cite{chen2024transunet}. By combining transformer based global context modeling with convolution based decoding, UNETR can preserve the advantages of U-Net while improving the ability to model complex spatial relationships. This makes UNETR suitable for volumetric medical segmentation tasks, especially when accurate segmentation depends on understanding broader anatomical context.

\renewcommand{\tabularxcolumn}[1]{m{#1}}
\newcolumntype{Y}{>{\centering\arraybackslash}X}

\begin{table*}[t]
    \centering
    \caption{Model Metrics Comparison on BraTS 2023 Dataset}
    \vspace{-0.2cm}
    \label{tab:model_comparison_brats2023}
    \renewcommand{\arraystretch}{1.25}
    \begin{tabularx}{\textwidth}{lYYYYYYYY}
        \toprule
        \textbf{Model} 
        & \makecell{\textbf{Training}\\\textbf{Loss}} 
        & \textbf{Precision} 
        & \textbf{Recall} 
        & \makecell{\textbf{F1} \textbf{Score}} 
        & \makecell{\textbf{Dice Mean}\\\textbf{(WT)}} 
        & \makecell{\textbf{Dice Mean}\\\textbf{(ET)}} 
        & \makecell{\textbf{Dice Mean}\\\textbf{(TC)}} 
        & \makecell{\textbf{Avg Dice}\\\textbf{Mean}} \\
        \midrule        
        Attention U-Net     & 0.3421 & 0.9929 & 0.9922 & 0.9925 & 0.6657 & 0.7392 & 0.5519 & 0.6522 \\
        U-Net 3D            & 0.1338 & 0.9876 & 0.9812 & 0.9834 & 0.6836 & 0.7164 & 0.6671 & 0.6890 \\
        Res U-Net           & 0.1336 & 0.9946 & 0.9948 & 0.9947 & 0.7315 & 0.7774 & 0.6934 & 0.7341 \\
        UNETR               & 0.1981 & 0.9966 & 0.9967 & 0.9967 & 0.8449 & 0.8720 & 0.8451 & 0.8540 \\
         \textbf{Swin UNETR} & 0.1295 & 0.9974 & 0.9974 & 0.9974 & 0.8966 & 0.9088 & 0.8841 & \textbf{0.8965} \\
        \bottomrule
    \end{tabularx}
\end{table*}

\subsection{Swin UNETR}
Swin UNETR further extends transformer based medical image segmentation by incorporating Swin Transformer blocks into the U-Net framework \cite{hatamizadeh2021swin}. Compared with standard transformer models, which may require high computational cost when processing large volumetric images, Swin Transformer uses window based self attention and shifted window self attention to improve efficiency while preserving contextual learning capability \cite{hatamizadeh2021swin}. As shown in Figure \ref{fig:BTSSwinUNETR}, Swin UNETR adopts a hierarchical transformer encoder and a U-Net style decoder with residual convolutional blocks and skip connections.

In Swin UNETR, the encoder extracts hierarchical representations from the input volume \cite{hatamizadeh2021swin}. Window based self attention models local relationships within each window, while shifted window self attention allows information exchange across neighboring windows. This mechanism enables the model to capture both local structure and broader contextual relationships. The hierarchical encoder design also provides multi scale feature representations, which are important for segmenting medical structures with different sizes and shapes \cite{hatamizadeh2021swin}.

The decoder of Swin UNETR follows a U-Net style reconstruction process \cite{cao2022swin}. It uses convolutional decoding blocks and skip connections to fuse multi scale encoder features and recover high resolution segmentation outputs. This combination allows Swin UNETR to preserve local spatial details while benefiting from transformer based global context modeling.

Swin UNETR is particularly effective for complex medical image segmentation tasks because it integrates three important capabilities: local feature extraction, long range dependency modeling, and multi scale representation learning \cite{heidari2023hiformer}. For brain tumor segmentation, these properties are useful because tumor regions can be irregular, heterogeneous, and spatially complex \cite{hatamizadeh2021swin}. For retinal vessel segmentation, the hierarchical representation can also help distinguish fine vessel structures from noisy background regions \cite{li2024td}. Therefore, Swin UNETR provides a strong architecture for evaluating the benefits of transformer based U-Net variants in medical image segmentation.

\section{Performance Evaluation}
\label{Performance Evaluation}
This section presents the experimental setup and performance evaluation of the selected U-Net variants on brain MRI segmentation and retinal vessel segmentation tasks. The evaluation includes dataset descriptions, preprocessing procedures, quantitative results, and comparative analysis. The goal is to evaluate the segmentation accuracy, training behavior, and generalization ability of each model under a consistent experimental framework.

\subsection{Datasets}
Two benchmark medical image segmentation datasets were used in this study: BraTS 2023 and DRIVE. These datasets were selected because they represent two different segmentation scenarios. BraTS 2023 \cite{brats2023} focuses on three dimensional brain tumor segmentation from multimodal MRI scans, while DRIVE focuses on two dimensional retinal vessel segmentation from fundus images \cite{staal2004drive}.

The BraTS 2023 dataset \cite{brats2023} was used for brain tumor segmentation. It contains multimodal MRI scans with expert annotated tumor labels. Each case includes multiple MRI modalities, such as T1, contrast enhanced T1, T2, and FLAIR, which provide complementary information for identifying tumor regions. The segmentation task includes three tumor subregions: whole tumor, enhancing tumor, and tumor core. In this study, 600 images were used for training and 200 images were used for validation.

The DRIVE dataset \cite{staal2004drive} was used for retinal vessel segmentation. It contains color fundus images and manually annotated vessel masks. This dataset is widely used to evaluate vessel extraction methods because retinal vessels have thin, branching, and low contrast structures. The original DRIVE dataset contains 20 images for training and 20 images for validation. Because of the limited dataset size, data augmentation was applied to increase the number of training samples. After augmentation, the training set was expanded to 4000 images, and 64 images were used for validation.

\subsection{Preprocessing and Data Augmentation}
For the BraTS 2023 dataset \cite{brats2023}, preprocessing was performed to standardize the MRI inputs before training. First, noninformative black regions around the MRI volumes were removed to reduce unnecessary background information. Second, image intensities were normalized to the range [0, 1], which helps reduce intensity variation across scans and improves training stability. Third, each MRI volume and its corresponding segmentation label were padded or cropped to obtain a consistent input size for model training.

For the DRIVE dataset \cite{staal2004drive}, fundus images and their corresponding vessel masks were normalized and resized to a consistent input resolution. Since DRIVE contains a small number of original training images, data augmentation was applied to reduce overfitting and improve generalization. The augmentation process increased the diversity of the training set by generating transformed versions of the original images. The applied augmentation operations included rotation, flipping, cropping, and intensity adjustment. These transformations help simulate variations in retinal image appearance and improve model robustness.

All models were trained for a maximum of 100 epochs, and early stopping was applied when validation performance did not improve for three consecutive epochs.

\subsection{Segmentation Performance Across U-Net Variants}
The segmentation performance of each model was evaluated using quantitative metrics and training behavior. For BraTS 2023, Dice scores were calculated for the whole tumor, enhancing tumor, and tumor core regions, and the average Dice score was used to summarize overall segmentation performance. For DRIVE, the Dice score was calculated by comparing the predicted retinal vessel mask with the manual vessel annotation. Since DRIVE is a two dimensional retinal image dataset, model inputs were adapted to the two dimensional setting while maintaining the same evaluation protocol across all compared architectures. Training loss curves were used to analyze convergence behavior, while ROC curves and AUC values were used to compare discrimination capability across different threshold settings.

\renewcommand{\tabularxcolumn}[1]{m{#1}}

\begin{table*}[t]
    \centering
    \caption{Model Metrics Comparison on DRIVE Dataset}
    \label{tab:model_comparison_drive}
    \renewcommand{\arraystretch}{1.25}
    \begin{tabularx}{\textwidth}{lYYYYYY}
        \toprule
        \textbf{Methods} 
        & \makecell{\textbf{Training} \textbf{Loss}} 
        & \textbf{Precision} 
        & \textbf{Recall} 
        & \makecell{\textbf{F1} \textbf{Score}} 
        & \textbf{Accuracy} 
        & \makecell{\textbf{Dice} \textbf{Score}} \\
        \midrule        
        U-Net 3D            & 0.1030 & 0.9456 & 0.8572 & 0.8806 & 0.8572 & 0.5967 \\
        Attention U-Net     & 0.2534 & 0.9639 & 0.9563 & 0.9587 & 0.9563 & 0.7855 \\
        UNETR               & 0.2428 & 0.9652 & 0.9621 & 0.9633 & 0.9621 & 0.8015 \\
        Res U-Net           & 0.1719 & 0.9662 & 0.9632 & 0.9643 & 0.9632 & 0.8069 \\
        \textbf{Swin UNETR} & 0.2077 & 0.9671 & 0.9621 & 0.9637 & 0.9621 & \textbf{0.8078} \\
        \bottomrule
    \end{tabularx}
\end{table*}

\subsubsection{BraTS 2023: Brain Tumor Segmentation Results}
Table \ref{tab:model_comparison_brats2023} presents the performance comparison of the five U-Net variants on the BraTS 2023 dataset. The evaluated models include Swin UNETR, UNETR, Residual U-Net, 3D U-Net, and Attention U-Net. The reported metrics include training loss, precision, recall, F1 score, Dice score for whole tumor (WT), Dice score for enhancing tumor (ET), Dice score for tumor core (TC), and the average Dice score.

Among all evaluated models, Swin UNETR achieved the strongest overall performance, with a training loss of 0.1295 and the highest average Dice score of 0.8965. It also achieved strong Dice scores across all three tumor subregions, with 0.8966 for WT, 0.9088 for ET, and 0.8841 for TC. UNETR obtained the second highest average Dice score of 0.8540, suggesting that transformer based global context modeling is beneficial for volumetric brain tumor segmentation. Residual U-Net achieved an average Dice score of 0.7341, while 3D U-Net and Attention U-Net obtained lower average Dice scores of 0.6890 and 0.6522, respectively.

\begin{figure}
    \centering
    \includegraphics[width=0.8\linewidth]{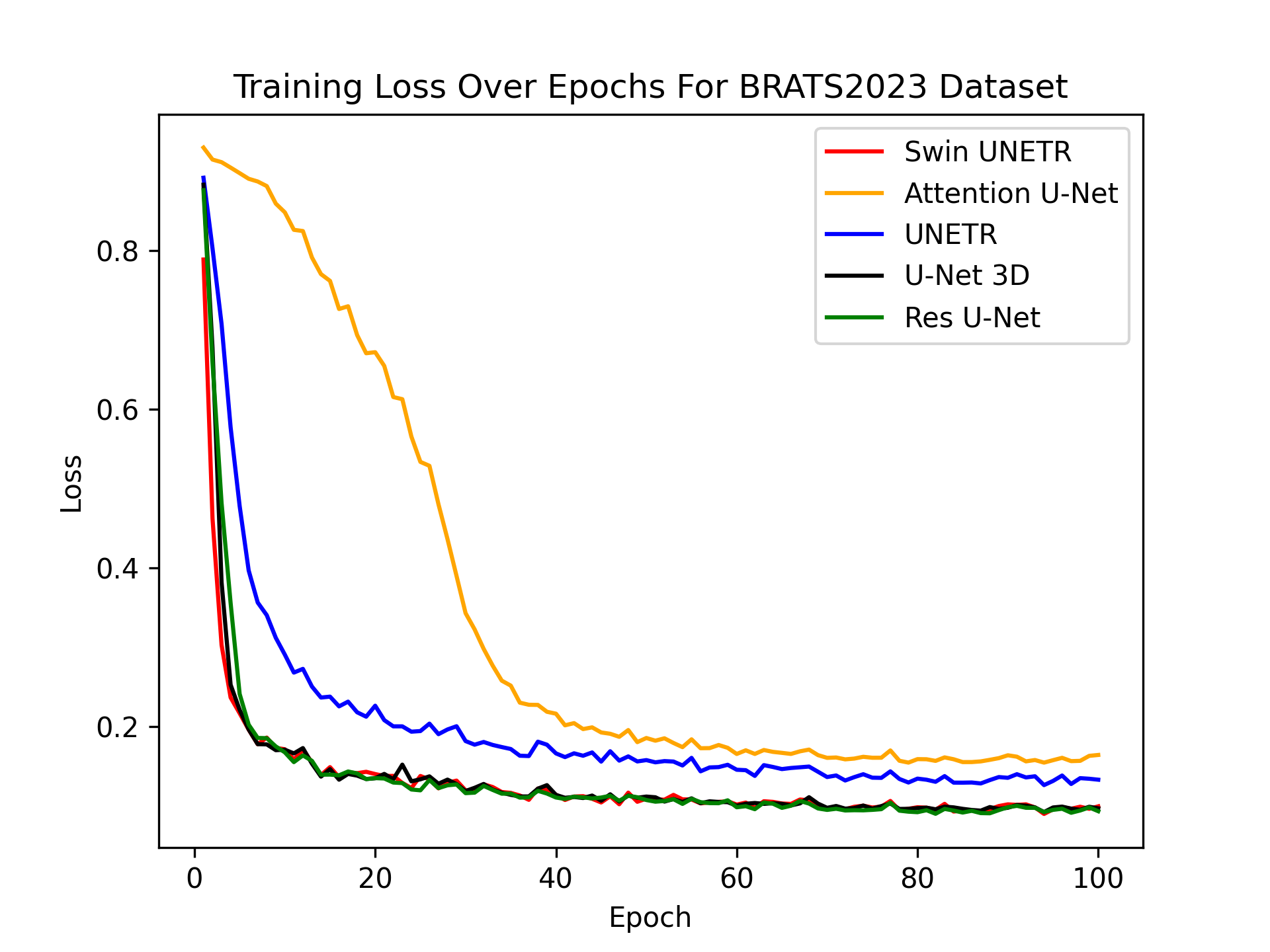}
    \vspace{-0.4cm}
    \caption{Training loss curves of the evaluated U-Net based models.}
    \label{figure:training-loss-over-epochs-brats2023}
\end{figure}

\begin{figure}
    \centering
    \includegraphics[width=0.8\linewidth]{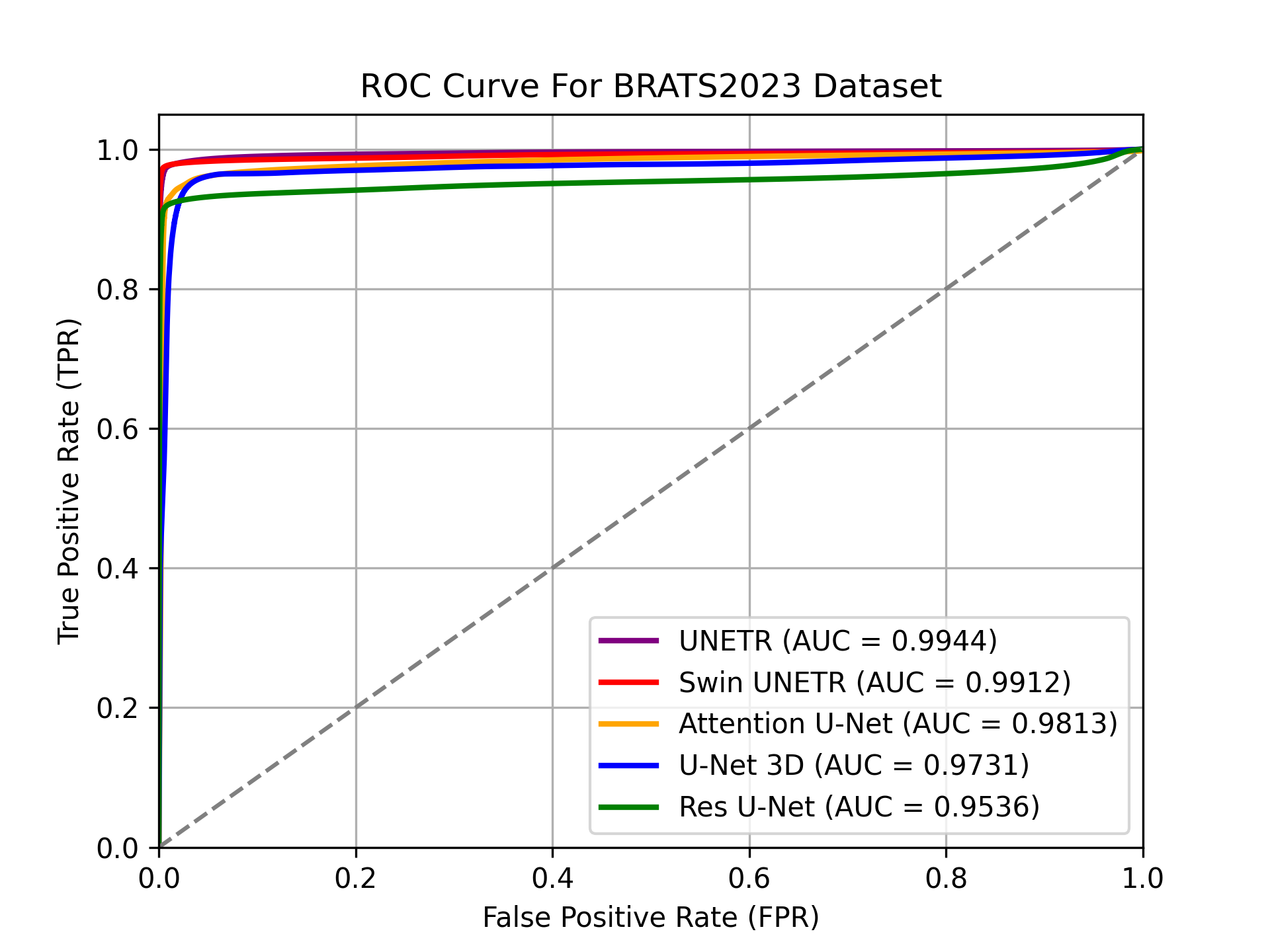}
    \vspace{-0.4cm}
    \caption{ROC curves of the evaluated U-Net based models.}
    \label{figure:roc-brats2023}
\end{figure}

These results indicate that transformer based U-Net variants, especially Swin UNETR, are effective for segmenting complex tumor structures in multimodal MRI scans. The strong performance of Swin UNETR may be attributed to its hierarchical transformer encoder and shifted window self attention mechanism, which allow the model to capture both local spatial details and broader contextual relationships. This capability is important for brain tumor segmentation because tumor regions often have irregular shapes, heterogeneous appearances, and ambiguous boundaries.

Figure \ref{figure:training-loss-over-epochs-brats2023} shows the training loss curves of the five models on the BraTS 2023 dataset. All models show a rapid decrease in training loss during the early training stage, indicating effective learning. Swin UNETR, Residual U-Net, and 3D U-Net reach relatively low and stable training losses, while Attention U-Net converges more slowly and remains higher than the other models. Figure \ref{figure:roc-brats2023} presents the ROC curves for the BraTS 2023 dataset. Swin UNETR and UNETR demonstrate stronger discrimination capability than the convolution based models, further supporting the advantage of transformer based architectures for complex volumetric tumor segmentation.

\subsubsection{DRIVE: Retinal Vessel Segmentation Results}
Table \ref{tab:model_comparison_drive} presents the performance comparison of the five U-Net variants on the DRIVE dataset. The evaluated metrics include training loss, precision, recall, F1 score, accuracy, and Dice score. Unlike BraTS 2023, DRIVE is a two dimensional retinal vessel segmentation dataset with thin and branching vascular structures. Therefore, this task emphasizes fine boundary preservation and robust vessel detection.

Swin UNETR achieved the highest Dice score of 0.8078 on DRIVE, followed closely by Residual U-Net with 0.8069 and UNETR with 0.8015. These results show that Swin UNETR remains effective for retinal vessel segmentation and that Residual U-Net also provides competitive performance. The strong performance of Residual U-Net suggests that residual feature learning is useful for preserving local structural details, which is important for segmenting thin and low contrast vessels. Attention U-Net achieved a Dice score of 0.7855, showing moderate performance compared with the other models.

\begin{figure}
    \centering
    \includegraphics[width=0.8\linewidth]{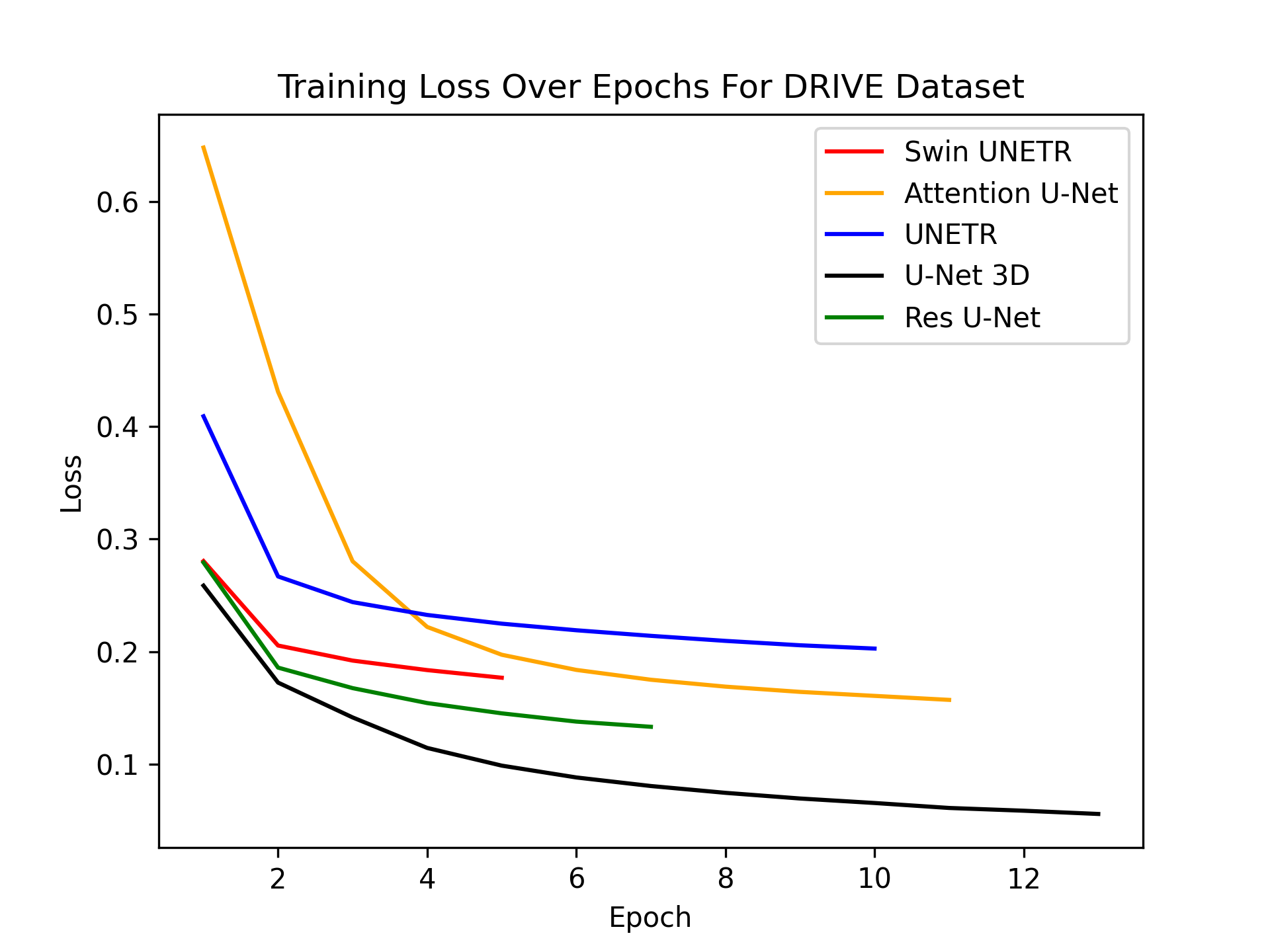}
    \vspace{-0.4cm}
    \caption{Training loss curves of the evaluated U-Net based models.}
    \label{figure:training_loss_over_epochs_drive}
\end{figure}

Although 3D U-Net achieved the lowest training loss of 0.1030, it produced the lowest Dice score of 0.5967. This result suggests that a lower training loss does not necessarily lead to better segmentation performance, especially when the architecture is less suitable for the target data structure. Since DRIVE contains two dimensional retinal images, models designed to preserve fine two dimensional vessel boundaries or capture broader contextual information can generalize better than a volumetric convolution based model adapted to this setting.

\begin{figure}
    \centering
    \includegraphics[width=0.8\linewidth]{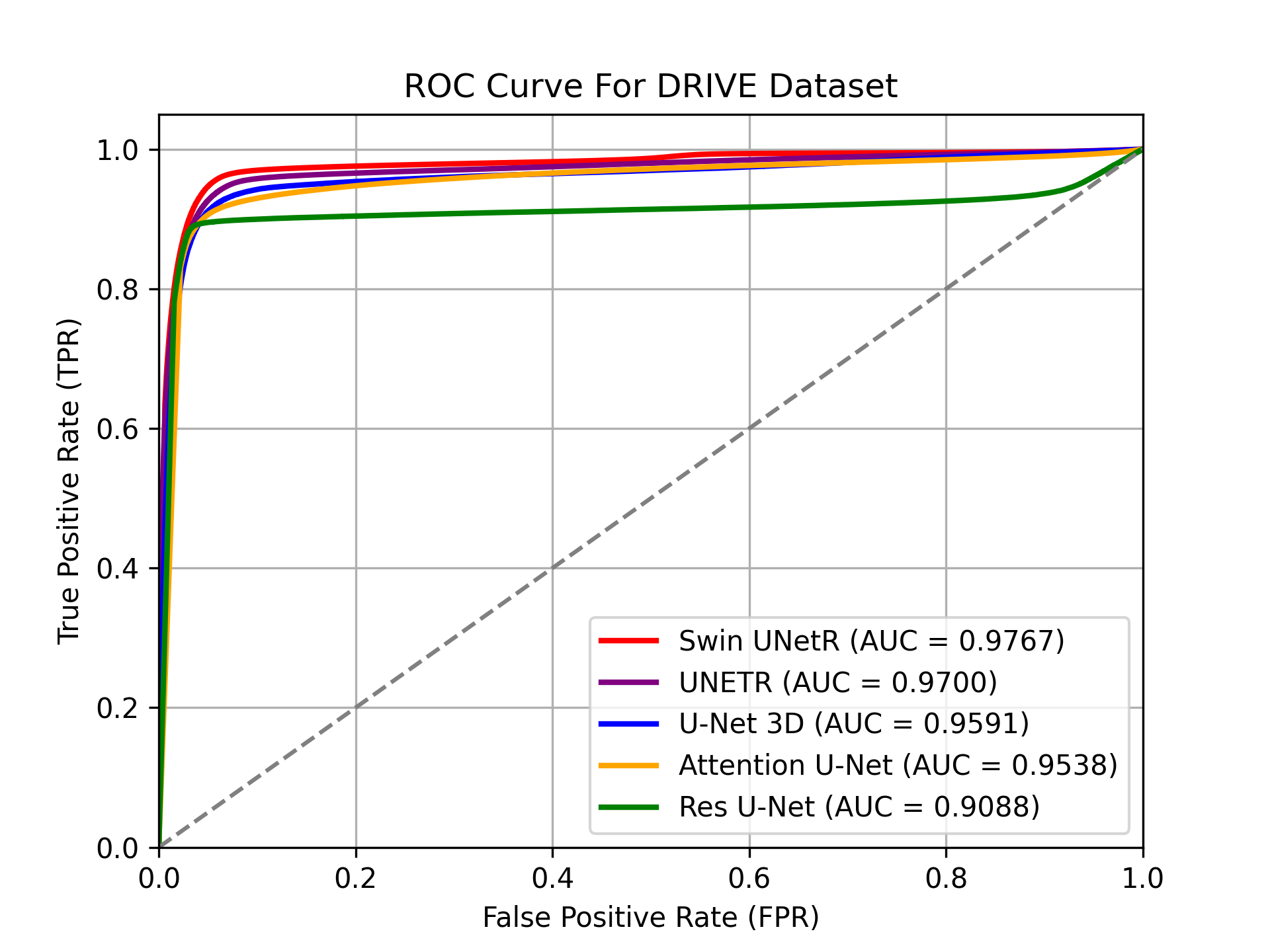}
    \vspace{-0.4cm}
    \caption{ROC curves of the evaluated U-Net based models.}
    \label{figure:roc-drive}
\end{figure}

Figure \ref{figure:training_loss_over_epochs_drive} shows the training loss curves on the DRIVE dataset. All models show decreasing training loss, indicating that the models learned from the training data. However, the difference between training loss and Dice score shows that convergence behavior alone is not sufficient to evaluate segmentation quality. Figure \ref{figure:roc-drive} presents the ROC curves on DRIVE. Swin UNETR, UNETR, and Residual U-Net show stronger discrimination performance than Attention U-Net and 3D U-Net, which is consistent with the Dice score results reported in Table \ref{tab:model_comparison_drive}.

\subsection{Discussion}
\vspace{-0.1cm}
The experimental results show that transformer based U-Net variants generally achieve stronger segmentation performance than purely convolution based models across the two evaluated datasets. On the BraTS 2023 dataset, Swin UNETR achieved the highest average Dice score among all compared models. This result suggests that hierarchical transformer representations and shifted window self attention are effective for brain tumor segmentation, where tumor regions may have irregular shapes, heterogeneous intensity patterns, and complex spatial relationships. By capturing both local features and broader contextual information, Swin UNETR provides more accurate segmentation of tumor subregions than the baseline U-Net 3D and other convolution based variants.

UNETR also achieved strong performance on BraTS 2023, which further confirms the benefit of transformer based global context modeling in volumetric medical image segmentation. However, Swin UNETR outperformed UNETR, which may be attributed to its hierarchical design and shifted window attention mechanism. These mechanisms allow Swin UNETR to model contextual relationships more efficiently while preserving multi scale feature representations. In contrast, U-Net 3D and Attention U-Net obtained lower average Dice scores, indicating that local convolutional operations and attention gates alone may be insufficient for capturing the complex anatomical and pathological variations present in brain tumor MRI data.

On the DRIVE dataset, Swin UNETR again achieved the highest Dice score, followed closely by Res U-Net and UNETR. The small performance difference among these models suggests that retinal vessel segmentation benefits from both contextual modeling and strong local feature extraction. Retinal vessels are thin, branching structures, and accurate segmentation requires preserving fine boundary details. Res U-Net performs competitively because residual learning helps maintain stable feature propagation and supports the detection of fine vessel patterns. Although U-Net 3D achieved the lowest training loss on DRIVE, its Dice score was substantially lower than those of the other models. This result suggests that lower training loss does not necessarily indicate better segmentation performance, especially when the model does not generalize well to fine two dimensional vessel structures.

Overall, the results indicate that model selection should depend on the segmentation task. For complex volumetric segmentation tasks such as brain tumor segmentation, transformer based architectures, especially Swin UNETR, are more suitable because they can capture long range contextual information and multi scale representations. For fine structure segmentation tasks such as retinal vessel extraction, both transformer based models and residual U-Net variants can provide strong performance. These findings suggest that future work should explore hybrid architectures that combine transformer based contextual modeling with convolutional or residual modules designed to preserve fine local details.

\section{Conclusion}
\label{Conclusion}
This paper presented a comparative study of U-Net based architectures for medical image segmentation on the BraTS 2023 and DRIVE datasets. The evaluated models included U-Net 3D, Residual U-Net, Attention U-Net, UNETR, and Swin UNETR, covering convolution based, residual, attention based, and transformer based segmentation approaches. The experimental results show that Swin UNETR achieved the best overall performance on both datasets, with the highest Dice scores for brain tumor segmentation and retinal vessel segmentation. These results suggest that hierarchical transformer representations and shifted window self attention are effective for capturing both local spatial details and long range contextual information. Res U-Net also produced competitive results on DRIVE, indicating that residual feature learning remains useful for fine structure segmentation tasks. Overall, the findings show that model selection should depend on the complexity and structure of the medical imaging task. Transformer based U-Net variants are suitable for complex segmentation tasks that require global contextual modeling, while residual models can remain effective for tasks that require strong local feature preservation. Future work will explore additional U-Net variants, hybrid architectures, customized loss functions, and larger clinical datasets to further improve segmentation robustness and generalization.


\bibliographystyle{IEEEtran}
\bibliography{refs}

\end{document}